\documentclass[letterpaper, 10 pt, conference]{ieeeconf}  

\IEEEoverridecommandlockouts                              

\title{\LARGE \bf
Learning  to  Fill  the  Seam by Vision: Sub-millimeter Peg-in-hole \\ on Unseen Shapes in Real World
}

\author{Liang Xie, Hongxiang Yu, Yinghao Zhao, Haodong Zhang, Zhongxiang Zhou, \\Minhang Wang, Yue Wang, Rong Xiong
\thanks{This work was supported by the National Nature Science Foundation of China under Grant 62173293. (\emph{Corresponding auther: Yue Wang and Rong Xiong.})}
\thanks{Liang Xie, Hongxiang Yu, Yinghao Zhao, Haodong Zhang, Zhongxiang Zhou, Yue Wang, Rong Xiong are with the State Key Laboratory of Industrial Control Technology and Institute of Cyber-Systems and Control, Zhejiang University, Zhejiang, China.  Minhang Wang is with the Application Innovate Lab, Huawei Incorporated Company, P.R. China. Yue Wang is the corresponding author {\tt\small wangyue@iipc.zju.edu.cn}. Rong Xiong is the co-corresponding auther {\tt\small rxiong@zju.edu.cn}.} }

\usepackage{multirow}
\usepackage{bbding}
\usepackage{amsmath} 
\usepackage{graphicx}
\usepackage{amssymb}
\usepackage[colorlinks, linkcolor=blue, citecolor=blue]{hyperref}
\usepackage{footnote}
\usepackage[flushleft]{threeparttable}
\usepackage{soul}
\usepackage{color, xcolor}
\usepackage{url}
\usepackage{verbatim}
\soulregister{\cite}7
\soulregister{\ref}7

\begin{document}

\maketitle
\thispagestyle{empty}
\pagestyle{empty}

\begin{abstract}
In the peg insertion task,  human pays attention to the seam between the peg and the hole and tries to fill it continuously with visual feedback. By imitating the human's behavior, we design architectures with position and orientation estimators based on the seam representation for pose alignment, which proves to be general to the unseen peg geometries. By putting the estimators into the closed-loop control with reinforcement learning, we further achieve higher or comparable success rate, efficiency, and robustness compared with the baseline methods. The policy is trained totally in simulation without any manual intervention. To achieve sim-to-real, a learnable segmentation module with automatic data collecting and labeling can be easily trained to decouple the perception and the policy, which helps the model trained in simulation quickly adapting to the real world with negligible effort. Results are presented in simulation and on a physical robot. Code, videos, and supplemental material are available at \url{https://github.com/xieliang555/SFN.git}
\end{abstract}

\section{Introduction}
Peg-in-hole is a routine skill in human's daily life and industrial tasks. such as placing the key into the lock, plugging  the  charger  into  the  outlet  and  inserting  the  bolt into the nut. Conventional methods such as random search, spiral search and force servo are widely used due to the simplicity to achieve. But the insertion accuracy and efficiency decrease significantly as the initial pose uncertainty becomes larger. Vision guided peg-in-hole is a promising approach to deal with the scenarios with large initial error but still remains an open challenge.

Conventional white-box vision methods using position-based visual servo where a perception model is trained to perform pose estimation with position control as the policy. The method can be transferred to the new scenarios easily by retraining the perception model. But it struggles to achieve high levels of precision due to the accumulation of the pose estimation error, the robot-camera calibration error and the robot positioning error. Image error based visual servo can achieve higher success rate for insertion with the end-to-end control. But for each type of feature error such as point error, line error or area error,  not only the perception model need to be retrained but also the policy need to be re-designed. 

Learning driven black-box methods for vision peg-in-hole is similar to the image error based visual servo. Both learn the policy with the image as input and the robot command as output. But the learning based policy is parameterized with neural network without the need to manually design. However, it is hard for the method to be general to the unseen hole shapes. We consider there is no explicit division between the perception and the policy which prevents the generalization. Recently, \cite{c3} explicitly separates the representation and the policy by firstly designing the autoencoder which encodes multi-sensory data into the latent vector as the feature representation, then the RL agent is introduced as the policy to take as input the feature representation and output actions. The learned representation and policy have the ability to be general to the unseen shapes, but the system achieves precision of 2mm, which may not meet the requirements of modern autonomous robots as mentioned in \cite{c2} . In addition, they achieve sim-to-real by retraining the feature representation model and the policy model on physical robots, which is unsafe and time-consuming.

In this paper, inspired by the behavior that human look at the seam, and try to fill the seam by aligning the peg position and orientation to the hole gradually with visual feedback, we propose the Seam Filling Net (SFN) to imitate the process for the 4-DoF peg-in-hole, which is a common task in our daily routines \cite{c4}\cite{c9}. The proposed SFN is an intermediate gray-box method, borrowing ideas from both the conventional image-based servo pipeline and learning-based policy techniques. The SFN consists of three modules as shown in Fig. \ref{framework}: 1) a segmentation module that performs seam-peg segmentation which helps the policy trained in the simulation adapt to the real world insertions. 2) a position alignment module that outputs the 2D heatmap with position information. 3) an orientation alignment module that outputs the 1D heatmap with orientation information. By introducing the semantic representation of the seam , the proposed SFN explicitly separates the perception and the control with the segmentation module, and formulates the task as a data driven seam feature based visual servo task. With the seam as the intermediate representation, we design architectures consisted of explicit position and orientation estimators to add inductive bias into the policy network, which is expected to help improve generalization over unseen shapes. As a result, our proposed system can realize sub-mm insertions in real world while is general to unseen shapes without extra data or manual finetune.
Moreover, we simplify the sim-to-real process where we only need to retrain the seam segmentation module in real world. The training enjoys efficient and safe supervised learning, where contact-free manipulations between the peg and the hole are performed for data collection. The proposed sim-to-real pipeline distinguishes from the common practice for reinforcement learning, which conducts inefficient data sampling in real world with the potential risk of collision. To further accelerate the sim-to-real adaptation, an automatic data annotation pipeline for the seam segmentation is proposed.

We validate the system on insertion tasks in real world where progress is achieved with high efficiency, sub-mm precision and ability to be general to the unseen shapes under the accumulation of the segmentation error and the camera calibration error. The contributions are as follows.
\begin{itemize}
\item A framework imitating the behavior of human for peg-in-hole by filling the seam, where the seam segmentation serves as the intermediate representation to separate the perception and the control.

\item A seam filling policy that is general to unseen shapes with the designed SFN architecture, then using RL to achieve robust insertion with sequential decision.

\item A fast real world segmentation technique using efficient supervised learning to achieve sim-to-real, with automatic data collection and annotation.

\item Evaluation of the proposed method compared to baseline methods in both simulation and real world, which demonstrates higher success rate and efficiency, better generalization over unseen shapes and robustness to vision occlusion.
\end{itemize}

\section{Related Work}

\subsection{Peg-in-Hole Insertion}

Tremendous progress has been made to solve the peg-in-hole task with visual or force feedback. \cite{c1} tries to achieve fast cylinder peg-in-hole within 5s under large initial error by continuous visual servoing,  but the method only applies to the cylinder peg and is not general to other geometries. \cite{c2} combines the state-of-the-art object tracking algorithm and the passively adaptive mechanical hardware to complete precision 6-DoF insertion tasks. However, due to the complex in-hand manipulation, it is time-consuming to complete the whole insertion. Other force-based methods \cite{c4}\cite{c6}\cite{c5} have the natural advantages to be general over various geometries and achieve very high precision tasks, where the required precision even exceeds that of the robot. However, those methods usually assume a small initial error, typically less than 2mm, to insure insertion accuracy and efficiency, and the training process is usually performed in the real world due to the lack of fidelity of contact simulation. In this paper, our proposed method owes the strengths of both the vision-based and force-based methods, and makes up their shortcomings. We achieve the fast peg-in-hole insertion with unseen shape generalization by taking the visual images as feedback. On top of that, we try to explore the precision boundaries of the vision guided peg-in-hole tasks with only one RGB camera and extend the application scope to a higher precision level with sub-$mm$ tolerance.


\subsection{Deep Learning for Assembly}
 In recent years, deep learning has received a lot of attention in the community of robotic manipulation, where rapid progress has been made, especially for the robotic assembly by various deep learning techniques. Such as 6D pose estimation \cite{c7} and tracking \cite{c2}, object detection \cite{c8}, key point estimation \cite{c1}\cite{c10}, representation learning \cite{c3}\cite{c9} and deep recurrent neural network \cite{c11}. However, most of the methods either need to collect large amount of real world data for training, or to perform domain randomization to transfer the model from simulation to the highly structured environments, which is time-consuming and task-specific.
 
 Another line of work focuses on learning policies for assembly tasks by deep reinforcement learning (DRL)\cite{c4}\cite{c6}\cite{c5}\cite{c17}\cite{c15}\cite{c16}\cite{c18}.  Those methods have been proved to achieve successful insertions in the real world. However, DRL relies on large amount of random exploration with the environment for policy learning, which is inefficient especially in the real world. In addition, random exploration may be unsafe to the robot and the environment. \cite{c5}\cite{c15}\cite{c16} designed extra modules to improve sample efficiency but still did not fix the problem from the root. The proposed method introduces RL for policy learning which is trained totally in simulation and can be transferred to the real world with negligible effort.

\section{Seam Filling Net}
\begin{figure*}[thpb]
\centering
\includegraphics[height=60mm,width=1\textwidth]{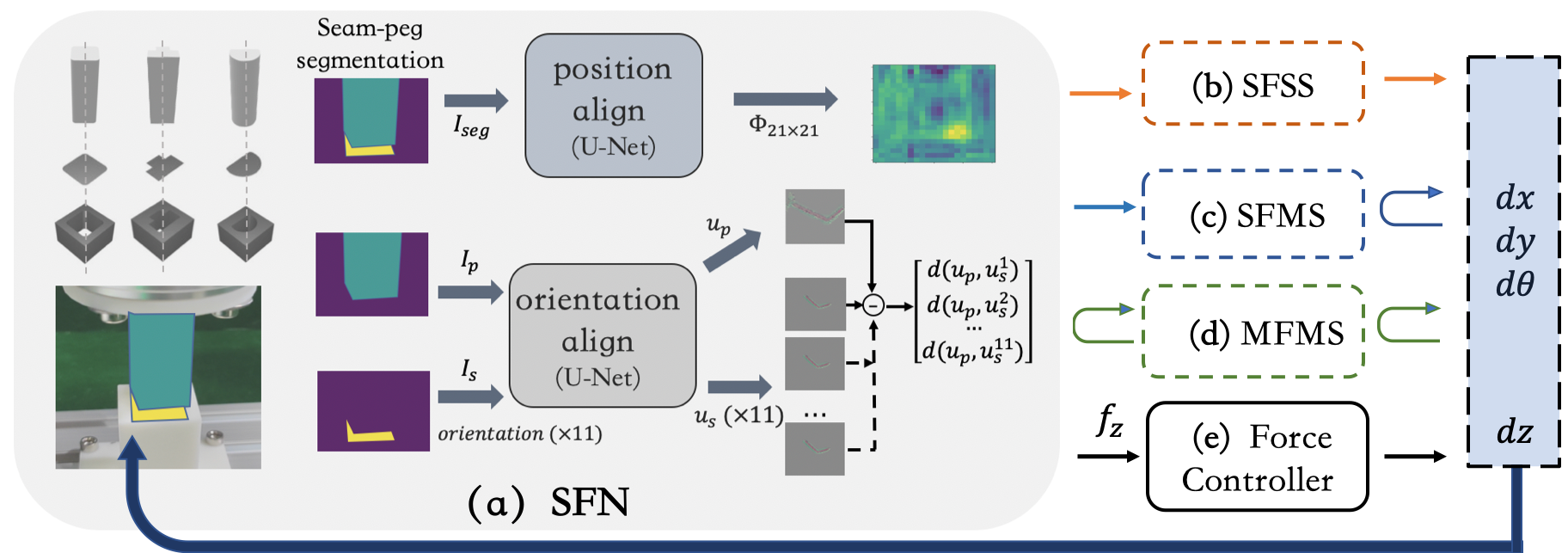}
\caption{\textbf{System framework with closed-loop control:} (a) The SFN consists of a seam segmentation module to obtain the peg-hole masks with the help of 3D models, a position module and a orientation module to produce heatmaps with position and orientation information by taking the seam representation as input. (b) The SFSS performs single-step action to obtain 3-DoF displacements by explicitly extracting the policy from the heatmaps. (c) The SFMS takes the single-frame heatmaps as perceptual inputs and tries to perform the optimal action by taking the future steps into consideration with RL. (d) The MFMS encodes the historical heatmaps of multi-frames with LSTM and tries to perform the optimal action with RL. (e) The peg and hole is kept constant contact in z-axis with force feedback from a F/T sensor, which is achieved by a PI based force controller.}
\label{framework}
\end{figure*}
The peg-in-hole task starts when both the peg and the hole are in the field of vision. The vision guided policy tries to eliminate the remaining 4-DoF error $dx, dy, dz, d \theta$. As the core of our method, the visual-based SFN generates the three-dimensional displacement of the robot's peg, which is defined over the translations $dx, dy$ and yaw angle $d\theta$ in the Cartesian space. And force control by a PI controller is applied to maintain constant force contact between the robot-peg and the hole in the z-axis as shown in Fig. \ref{framework}, which helps to eliminate vertical error $dz$.  We further introduce each module of the SFN in the following subsections.

\subsection{Single-frame Single-step SFN}

\textbf{Seam Segmentation:} The goal of the segmentation module is to obtain the segmented image $I_{seg}$, which contains three values, indicating for the peg, the seam and the background. The resolution of the image is 250$\times$200. The segmented image $I_{seg}$ then serves as the input to the following modules for both position and orientation alignments. In this way, the segmentation module decouples the perception and the policy, which helps the policy trained in simulation directly adapt to the real world insertions. Thanks to the efficient supervised learning, the segmentation module can be easily trained among simulation and real world.

\textbf{Position Alignment Module:} The position alignment module tries to focus on the relative position between the peg and the seam by making 2D-translate prediction from one-hot heatmaps. Let $dx^*$ and $dy^*$ be the ground truth of displacement along the X-axis and Y-axis respectively, $i$ and $j$ be the corresponding pixel coordinates. The $dx^*$ and $dy^*$ can be obtained from simulation by translating the displacements from world coordinates to pixel coordinates with a calibrated camera. Then the ground truth one-hot heatmap $\Phi^*$ with a pixel resolution of $n\times n$ can be defined as:
\begin{equation}
\Phi_{i,j}^* = 
\begin{cases}
1 & \text{if $dx^* = i-\frac{n}{2}$, $dy^*=j-\frac{n}{2}$}  \\ 
0 & \text{o.w.}
\end{cases}
\end{equation}

We choose the U-Net $f_\varphi$ as our backbone to predict the one-hot heatmap from the image $I_{seg}$ as Eq. \ref{seg}, where $\varphi$ is the learnable parameters. The U-Net outputs two channels, one of which predicts the highlights and the other the background.  A loss function is defined over the ground truth one-hot heatmap $\Phi^*$ and the predicted one-hot heatmaps $\Phi$ with binary cross entropy loss as in Eq. \ref{pos-loss}.
\begin{gather}
\Phi=f_{\varphi}(I_{seg}) \label{seg}\\
Loss_{\varphi}=-\sum(\Phi^*\cdot log\Phi + (1-\Phi^*)\cdot log(1-\Phi)) \label{pos-loss}
\end{gather}

\textbf{Orientation Alignment Module:} 
The orientation alignment module uses the U-Net $f_{\psi}$ as the backbone which consists of a two-stream Siamese network with shared parameters $\psi$. The first stream takes as input the segmented image $I_{p}$ which only contains the pixels of the peg and the background. The second stream takes as input the 11 rotated and segmented images $I_{s}$ which only contain the pixels of the seam and the background, with 5 equispaced clockwise orientations and 5 equispaced anticlockwise orientations. The rotation resolution is 2 degrees. Both streams output 3-channel feature with the same resolution of the input image, defined as $u_{p}, u_{s}$ respectively. The $11\times1$ heatmap produced by the orientation module is defined as:
\begin{gather}
    u_p = f_{\psi}(I_p)\\
    u_s = f_{\psi}(I_s)\\
    D_i = e^{-d(u_p, u_s^{i})}
    \vspace{-0.5cm}
\end{gather}
where $D_i$ is the $i-th$ element of the heatmap and $d$ is the Euclidean distance between the two feature maps. Among the 11 oriented images $I_{s}$, only the correct one $I^{p}_{s}$ with the ground truth orientation will match the image $I_{s}$ and the others $I^{n}_{s}$ will mismatch. To establish the orientation correspondences between the peg and the seam, the contrastive loss function is designed to encourage the paired peg-seam feature maps to match, while push the unpaired maps to be a feature distance margin 1 apart:
\begin{equation}
    Loss_{\psi} = \max\{D_n-D_p+1, 0\}
\end{equation} 
We define the single-frame single-step policy (SFSS) by explicitly calculating the displacements  from the position and orientation alignment modules as discussed above. The displacements $dx$, $dy$, $d\theta$ can be obtained by taking the maximum value of heatmaps produced by the position module and the orientation module respectively. The SFSS-SFN tries to learn a policy which is general to the unseen geometries by filling the seam. However, there remains several drawbacks under random initial error circumstances. There exist multiple solutions due to the vision occlusion, where the policy may get stuck with the disappearance of the seam in the vision field. Moreover, the insertion tends to fail when dealing with larger initial error due to incorrect displacement estimation in the open-loop control.  Despite that the single-step SFN can achieve the closed-loop control by recursively output the one-step-optimal action, the policy is essentially a single-step decision process without considering the future steps to make the optimal decision.

\subsection{Single-frame Multi-step SFN}
To overcome the drawbacks discussed above, we propose the single-frame multi-step SFN(SFMS) for robust peg-in-hole by formulating the task as the multi-step sequential learning problem with RL.

The peg-in-hole task can be modelled as a finite-horizon Markov decision process $\mathcal{M}$, with a state space $\mathcal{S}$, an action space $\mathcal{A}$, a reward function $r: \mathcal{S}\times{\mathcal{A}} \to \mathbb{R}$ and a discount factor $\gamma  \in (0,1]$. The RL agent observes the current state $s_t$ defined as:
\begin{equation}
    s_t = [\Phi, D]\label{eq6}
\end{equation}
where $\Phi$ and $D$ are the flattened heatmaps produced by the position alignment module and the orientation alignment module respectively. Then the policy $\pi$ generates an action $a_t$ to the robot control defined as:
\begin{equation}
    a_t = [dx, dy, d\theta] =\pi(s_t)
\end{equation}
A dense reward $r$ is obtained from the environment after each step of performing the robot command defined as:
\begin{equation}
r = 
\begin{cases}
1 -\alpha Loss_{\varphi} -\beta Loss_{\psi} & success  \\ 
-\frac{1}{k_{max}} -\alpha Loss_{\varphi} -\beta Loss_{\psi} & o.w.
\end{cases}
\end{equation}
where $k_{max}$ is the preset maximum steps of an episode. The reward is designed to encourage the policy $\pi$ to complete insertion with the minimum steps, and enforce the policy to learn the visual representations that help achieve robust insertion by punishing the loss function of both the orientation and the position modules. The episode finishes when reaching the predefined maximum steps or a successful insertion. 

We use Advantage Actor-Critic \cite{c19}, a modern policy gradient based RL algorithm, as the policy architecture. The loss function for RL can be found in the Appendix \cite{c22}.

\subsection{Multi-frame Multi-step SFN}
The SFMS makes decision by taking the future steps into account with RL to achieve robust insertion, which may still result in uncertainty with only one-step perception as input. By aggregating the temporal feature with LSTM networks,  we propose to introduce the multi-frame multi-step policy (MFMS) to address the one-step perception uncertainty.

We introduce the LSTM network $\mathcal{L}$ parameterized by $\upsilon$ to encode the historical sequential heatmaps produced by the position and the orientation modules. By stacking Eq. \ref{eq6} with the latest n-steps of heatmaps and encoding them with LSTM, the current state is redefined as: 
\begin{equation}
    s_t = \mathcal{L_{\upsilon}}([\Phi, D]_{t-n:t})
\end{equation}
The state is then took as input for the RL agent. 

\subsection{Efficient Labeling for Sim-to-Real}
We can achieve sim-to-real by only retraining the segmentation module in real world, by which the segmented image can be directly took as input for the policy trained in simulation for position and orientation alignment. To generate enough data for training the segmentation network, an automatic data collecting and labeling pipeline are designed. With the help of the 3D models of the peg-hole pairs as shown in Fig. \ref{framework}, the peg-hole masks can be obtained by rendering with known 3D models and poses in pyrender. The pose of the peg can be obtained by regarding the peg as the end-effector in the robot-peg system. And the hole pose relies on a human demonstration which can be obtained by performing an accurate insertion. By randomizing the robot-peg pose relative to the hole, large amount of image-mask pairs can be obtained for the segmentation training, which is fully automatic, time-efficient and low-cost.

\section{Experiment}
\textbf{Experiment setup:} We build a simulation environment in PyBullet \cite{c21} for model training. In simulation, we use the Franka Panda robot, a 7-DoF torque-controlled robot, of which the end-effector is connected with a peg with a fixed joint. An wrist F/T sensor is mounted to measure the contact force.  An camera is fixed at the space to look at both the peg and the hole Following \cite{c1}. We use Pyrender, a physically-based Python library for rendering and visualization, to get the segmentation masks . The initial position error ranges from -10mm to 10mm and the initial orientation error ranges from -10 degrees to 10 degrees. To test the geometry generalization of the policy, 4 seen peg shapes are designed for training and 10 unseen peg shapes for evaluation. The 10 unseen peg shapes are selected elaborately from daily insertion tasks which range from simple to complex, concave to convex, and square-corner to round-corner as shown in the Appendix \cite{c22}. The tolerances of the peg-hole pairs are around 1mm. By imitating the layout of the simulation, we evaluate the proposed method on a real platform, which consists of a 6-DoF UR5 robot, a Robotiq FT300 wrist F/T sensor, and a rigid L515 camera calibrated with respect to the robot base. We use the peg-hole pairs with around $0.6mm$ tolerance for all the real world experiments which is tighter than the simulation setup. We design the 3D mesh of the peg-hole pairs  in Blender, an open source for 3D modeling, with the clearance between 0.5mm to 0.7mm. The 3D print error is around 0.2mm. The evaluation of shape generalization and feature visualization can be found in the Appendix \cite{c22}.

\begin{table*}[t]
\begin{threeparttable}[t]
\caption{Comparisons with the Baseline Methods}
\label{table_2}
\centering
\setlength{\tabcolsep}{1.8pt}
\begin{tabular}{ccclccccccccccc}
\hline
 & \multicolumn{3}{c}{\textbf{Seen}} & \multicolumn{11}{c}{\textbf{Unseen peg geometries}} \\
\multirow{-2}{*}{\textbf{Methods}} & 1 & \multicolumn{2}{c}{2} & 1 & 2 & 3 & 4 & 5 & 6 & 7 & 8 & 9 & 10 & avg(\%) \\ \hline
\multicolumn{15}{c}{\textbf{5mm / success rate}} \\ \hline
spiral search* & 10/12 & \multicolumn{2}{c}{11/12} & {\color[HTML]{CB0000} 10/12} & {\color[HTML]{CB0000} 11/12} & {\color[HTML]{CB0000} 10/12} & {\color[HTML]{CB0000} 10/12} & {\color[HTML]{CB0000} 10/12} & {\color[HTML]{CB0000} 11/12} & 7/12 & {\color[HTML]{CB0000} 8/12} & {\color[HTML]{CB0000} 8/12} & {\color[HTML]{CB0000} 7/12} & {\color[HTML]{CB0000} 78.47} \\
E2E Force-RL* & 11/12 & \multicolumn{2}{c}{10/12} & 9/12 & 9/12 & 8/12 & 9/12 & 7/12 & 7/12 & {\color[HTML]{CB0000} 8/12} & 7/12 & 7/12 & 7/12 & {\color[HTML]{3531FF} 68.75} \\
E2E Vision-RL & {\color[HTML]{CB0000} 12/12} & \multicolumn{2}{c}{{\color[HTML]{CB0000} 12/12}} & $\times$ & $\times$ & $\times$ & $\times$ & $\times$ & $\times$ & $\times$ & $\times$ & $\times$ & $\times$ & $\times$ \\
\textbf{SFSS-SFN-re (ours)} & {\color[HTML]{3531FF} 11/12} & \multicolumn{2}{c}{10/12} & 9/12 & 8/12 & 8/12 & 7/12 & 8/12 & {\color[HTML]{3531FF} 9/12} & $\times$ & 6/12 & 5/12 & $\times$ & $\times$ \\
\textbf{SFMS-SFN (ours)} & 10/12 & \multicolumn{2}{c}{{\color[HTML]{3531FF} 11/12}} & 8/12 & {\color[HTML]{3531FF} 9/12} & 8/12 & {\color[HTML]{3531FF} 9/12} & {\color[HTML]{3531FF} 8/12} & 8/12 & 6/12 & 7/12 & {\color[HTML]{3531FF} 7/12} & 5/12 & 66.67 \\
\textbf{MFMS-SFN (ours)} & 11/12 & \multicolumn{2}{c}{11/12} & {\color[HTML]{3531FF} 9/12} & 8/12 & {\color[HTML]{3531FF} 9/12} & 8/12 & 7/12 & 7/12 & {\color[HTML]{3531FF} 7/12} & {\color[HTML]{3531FF} 7/12} & 6/12 & {\color[HTML]{3531FF} 6/12} & 66.67 \\ \hline
\multicolumn{15}{c}{\textbf{5mm / efficiency(s)}} \\ \hline
spiral search* & $10\pm 3.1$ & \multicolumn{2}{c}{$9\pm3.0$} & $10\pm3.2$ & $9\pm3.4$ & $11\pm3.2$ & $10\pm3.7$ & $11\pm3.5$ & $9\pm3.1$ & $11\pm3.8$ & $10\pm3.3$ & $10\pm3.5$ & $11\pm3.9$ & $10\pm3.5$ \\
E2E Force-RL* & {\color[HTML]{3531FF} $4\pm1.6$} & \multicolumn{2}{c}{$3\pm1.1$} & $4\pm0.9$ & $5\pm1.2$ & {\color[HTML]{CB0000} $3\pm0.9$} & {\color[HTML]{3531FF} $4\pm1.0$} & {\color[HTML]{CB0000} $3\pm1.0$} & $5\pm1.1$ & $11\pm2.3$ & $6\pm1.5$ & $7\pm1.5$ & $10\pm2.3$ & $5\pm2.5$ \\
E2E Vision-RL & $5\pm0.9$ & \multicolumn{2}{c}{{\color[HTML]{3531FF} $3\pm1.1$}} & $\times$ & $\times$ & $\times$ & $\times$ & $\times$ & $\times$ & $\times$ & $\times$ & $\times$ & $\times$ & $\times$ \\
\textbf{SFSS-SFN-re (ours)} & {\color[HTML]{CB0000} $4\pm1.4$} & \multicolumn{2}{c}{$3\pm1.2$} & $5\pm1.6$ & {\color[HTML]{3531FF} $3\pm0.6$} & $4\pm0.5$ & $4\pm1.7$ & $5\pm1.2$ & {\color[HTML]{3531FF} $4\pm1.5$} & $\times$ & {\color[HTML]{3531FF} $4\pm1.5$} & {\color[HTML]{3531FF} $5\pm1.3$} & $\times$ & $\times$ \\
\textbf{SFMS-SFN (ours)} & $5\pm1.9$ & \multicolumn{2}{c}{{\color[HTML]{CB0000} $3\pm0.9$}} & {\color[HTML]{3531FF} $4\pm0.4$} & {\color[HTML]{CB0000} $2\pm0.3$} & $5\pm2.1$ & {\color[HTML]{CB0000} $3\pm0.6$} & $4\pm3.1$ & {\color[HTML]{CB0000} $3\pm1.2$} & {\color[HTML]{CB0000} $6\pm2.5$} & $4\pm1.5$ & $5\pm1.6$ & {\color[HTML]{CB0000} $7\pm3.9$} & {\color[HTML]{CB0000} $4\pm1.3$} \\
\textbf{MFMS-SFN (ours)} & $5\pm0.7$ & \multicolumn{2}{c}{$4\pm1.1$} & {\color[HTML]{CB0000} $3\pm2.8$} & $5\pm3.3$ & {\color[HTML]{3531FF} $3\pm1.6$} & $5\pm3.2$ & {\color[HTML]{3531FF} $3\pm1.2$} & $5\pm2.3$ & {\color[HTML]{3531FF} $7\pm2.1$} & {\color[HTML]{CB0000} $4\pm1.1$} & {\color[HTML]{CB0000} $3\pm1.3$} & {\color[HTML]{3531FF} $9\pm3.3$} & {\color[HTML]{3531FF} $4\pm1.7$} \\ \hline
\multicolumn{15}{c}{\textbf{10mm / success rate}} \\ \hline
spiral search* & 5/12 & \multicolumn{2}{c}{6/12} & 6/12 & 7/12 & 6/12 & 5/12 & 5/12 & 4/12 & 1/12 & 3/12 & 3/12 & 3/12 & 37.5 \\
E2E Force-RL* & $\times$ & \multicolumn{2}{c}{$\times$} & $\times$ & $\times$ & $\times$ & $\times$ & $\times$ & $\times$ & $\times$ & $\times$ & $\times$ & $\times$ & $\times$ \\
E2E Vision-RL & {\color[HTML]{3531FF} 11/12} & \multicolumn{2}{c}{{\color[HTML]{CB0000} 12/12}} & $\times$ & $\times$ & $\times$ & $\times$ & $\times$ & $\times$ & $\times$ & $\times$ & $\times$ & $\times$ & $\times$ \\
\textbf{SFSS-SFN-re (ours)} & 10/12 & \multicolumn{2}{c}{10/12} & 8/12 & 9/12 & 7/12 & 7/12 & {\color[HTML]{CB0000} 8/12} & 6/12 & $\times$ & 4/12 & {\color[HTML]{3531FF} 5/12} & $\times$ & $\times$ \\
\textbf{SFMS-SFN (ours)} & 10/12 & \multicolumn{2}{c}{{\color[HTML]{3531FF} 11/12}} & {\color[HTML]{CB0000} 10/12} & {\color[HTML]{3531FF} 10/12} & {\color[HTML]{CB0000} 9/12} & {\color[HTML]{3531FF} 8/12} & 7/12 & {\color[HTML]{3531FF} 7/12} & {\color[HTML]{3531FF} 3/12} & {\color[HTML]{CB0000} 6/12} & 4/12 & {\color[HTML]{CB0000} 4/12} & {\color[HTML]{3531FF} 61.81} \\
\textbf{MFMS-SFN (ours)} & {\color[HTML]{CB0000} 11/12} & \multicolumn{2}{c}{11/12} & {\color[HTML]{3531FF} 9/12} & {\color[HTML]{CB0000} 10/12} & {\color[HTML]{3531FF} 8/12} & {\color[HTML]{CB0000} 9/12} & {\color[HTML]{3531FF} 8/12} & {\color[HTML]{CB0000} 7/12} & {\color[HTML]{CB0000} 4/12} & {\color[HTML]{3531FF} 5/12} & {\color[HTML]{CB0000} 6/12} & {\color[HTML]{3531FF} 3/12} & {\color[HTML]{CB0000} 63.19} \\ \hline
\multicolumn{15}{c}{\textbf{10mm / efficiency(s)}} \\ \hline
spiral search* & $23\pm4.5$ & \multicolumn{2}{c}{$25\pm4.6$} & $20\pm4.8$ & $20\pm4.4$ & $23\pm4.3$ & $22\pm4.4$ & $21\pm5.1$ & $24\pm4.5$ & $25\pm3.1$ & $22\pm4.5$ & $23\pm6.1$ & $25\pm4.3$ & $22\pm5.4$ \\
E2E Force-RL* & $\times$ & \multicolumn{2}{c}{$\times$} & $\times$ & $\times$ & $\times$ & $\times$ & $\times$ & $\times$ & $\times$ & $\times$ & $\times$ & $\times$ & $\times$ \\
E2E Vision-RL & $5\pm1.3$ & \multicolumn{2}{c}{$5\pm1.8$} & $\times$ & $\times$ & $\times$ & $\times$ & $\times$ & $\times$ & $\times$ & $\times$ & $\times$ & $\times$ & $\times$ \\
\textbf{SFSS-SFN-re (ours)} & $5\pm1.7$ & \multicolumn{2}{c}{{\color[HTML]{CB0000} $4\pm2.5$}} & {\color[HTML]{CB0000} $3\pm1.7$} & $5\pm2.5$ & $6\pm3.2$ & {\color[HTML]{CB0000} $4\pm0.6$} & {\color[HTML]{CB0000} $4\pm0.7$} & $6\pm1.9$ & $\times$ & $7\pm4.7$ & $8\pm3.7$ & $\times$ & $\times$ \\
\textbf{SFMS-SFN (ours)} & {\color[HTML]{3531FF} $4\pm1.3$} & \multicolumn{2}{c}{$6\pm2.1$} & {\color[HTML]{3531FF} $4\pm1.6$} & {\color[HTML]{3531FF} $4\pm1.9$} & {\color[HTML]{CB0000} $5\pm2.3$} & $6\pm1.5$ & {\color[HTML]{3531FF} $4\pm1.6$} & {\color[HTML]{3531FF} $5\pm1.7$} & {\color[HTML]{3531FF} $7\pm1.6$} & {\color[HTML]{3531FF} $4\pm1.0$} & {\color[HTML]{3531FF} $5\pm0.9$} & {\color[HTML]{CB0000} $7\pm1.2$} & {\color[HTML]{3531FF} $5\pm1.9$} \\
\textbf{MFMS-SFN (ours)} & {\color[HTML]{CB0000} $3\pm0.4$} & \multicolumn{2}{c}{{\color[HTML]{3531FF} $5\pm0.8$}} & $6\pm1.6$ & {\color[HTML]{CB0000} $4\pm0.5$} & {\color[HTML]{3531FF} $6\pm1.5$} & {\color[HTML]{3531FF} $5\pm1.1$} & $5\pm1.5$ & {\color[HTML]{CB0000} $3\pm1.6$} & {\color[HTML]{CB0000} $6\pm2.4$} & {\color[HTML]{CB0000} $3\pm1.4$} & {\color[HTML]{CB0000} $3\pm0.5$} & {\color[HTML]{3531FF} $7\pm2.6$} & {\color[HTML]{CB0000} $5\pm1.7$} \\ \hline
\end{tabular}
\begin{tablenotes}
\item[\scalebox{1.0}{$*$}] This superscript denotes the 3-DoF peg-in-hole.
\item[\scalebox{1.0}{$\times$}] This superscript denotes the method is not general to the corresponding geometry.
\item The red font values indicates the best results among the compared method.
\item The blue font values indicates the second best results among the compared method.
\end{tablenotes}
\end{threeparttable}
\vspace{-0.5cm}
\end{table*}

\begin{figure}[]
\centering
\includegraphics[height=60mm,width=0.49\textwidth]{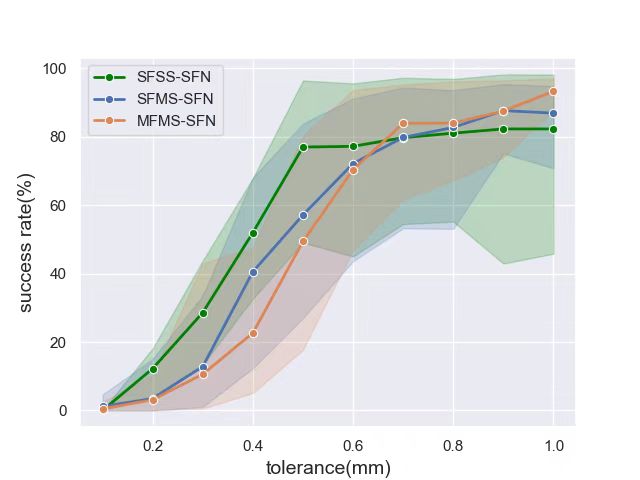}
\caption{Tolerance vs. Success rate }
\label{tolerance}
\vspace{-0.4cm}
\end{figure}

\textbf{Insertion precision:} We design simulation experiments on the relationship between the tolerance and the success rate to explore if sub-millimeter precision can be achieved with only one color camera. The tolerance of the peg-hole pairs ranges from 0.1 to 1mm with 10 steps. For each tolerance step, the insertion success rate is averaged over 4 seen peg shapes and 10 unseen shapes with each performing 12 insertion trials, which counts to 1680 trials in total. As shown in Fig. \ref{tolerance}, when the tolerance ranges from 0.6mm to 1mm, our proposed system still maintains a high insertion success rate for all the three methods. 


\subsection{Simulation Experiments}

\begin{figure}[]
\centering
\includegraphics[height=60mm,width=0.49\textwidth]{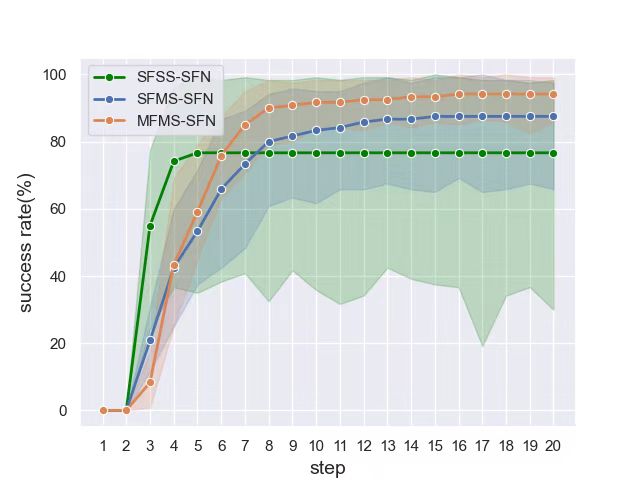}
\caption{Insertion steps vs. Success rate}
\label{step}
\vspace{-0.4cm}
\end{figure}

\textbf{Closed-loop control:} Simulation experiments on relationship between the insertion steps and the success rate are performed in this section to evaluate how closed-loop control influence the insertion accuracy. Similarly, for each of the 20 step, the success rate is averaged over 4 seen peg shapes and 10 unseen shapes with each performing 12 insertion trials, resulting in 2800 trials in total. As shown in Fig. \ref{step}, all the compared methods achieve rapid progress on success rate with the increasing steps, and the curves start to converge within 5 steps, which proves the inference that the closed-loop control helps improve success rate. However, the SFSS-SFN does not converge to the same accuracy as the other two methods, we infer that in the vision occlusion case, the policy gets stuck by recursively performing the step-optimal policy, where more steps can not help the policy to get out of the visual blind when the current estimation fails. In addition, our full system MFMS-SFN converges to the higher success rate than the SFMS-SFN by encoding the historical frames as the current state, which helps reduce the single-frame uncertainty when making decisions.

\subsection{Real World Insertions}

\textbf{Benchmark:} We compare the proposed system with five baseline alternatives in the real world platform on insertion efficiency, success rate and generalization to the unseen peg geometries. For all the learning-based methods, we select the 4 seen peg geometries to train the policy and evaluate it on the 10 unseen geometries as with the simulation setting.

\begin{itemize}
    \item \textbf{Spiral Search} \cite{c1}: Spiral search will move downward in the z-axis until a force limit is reached, indicating the peg is in contact with the hole. Then the peg moves outwards in a spiral while pressing against the hole-surface. When the peg moves over the hole within a success region, the force in z-axis will help press the peg to insert. Otherwise the peg will moves until exceeding the uncertainty boundary and fail.
    \item \textbf{E2E Force-RL} \cite{c6}: The E2E Force-RL learns the policy end-to-end, which outputs the 2D translate displacement $[dx, dy]$ in the horizon space by taking the 6D force/torque $[F_x, F_y, F_z, M_x, M_y, M_z]$ as input. Force control is performed in the z-axis to maintain a constant contact as with the spiral search. The sparse reward is given when the insertion successes within possibly the minimum steps as defined in \cite{c6}. The model trained in simulation is directly transferred to real world.
    \item \textbf{E2E Vision-RL} \cite{c20}: The E2E Vision-RL differs with the Force-RL by taking the raw RGB image as input instead of the 6D force/torque data. Here to make it fair, we use the segmented image as input as with our proposed method to achieve sim-to-real. The policy network, action space and reward function is defined the same as in \cite{c20}. 
    \item \textbf{SFSS-SFN-re:} The SFSS-SFN-re uses the SFSS-SFN as backbone and outputs multi-step actions by recursively performing the one-step policy. It is important to note that the "multi-step" here differs from the multi-step backbone in our proposed system. Specifically, during training, the single-step SFN only learns the one-step policy by supervised learning, which is a step-optimal policy, while the multi-step SFN learns multi-step policy in an episode by reinforcement learning, which is a trajectory-optimal policy. 
    \item \textbf{SFMS-SFN:} The SFMS-SFN takes the single-frame perception as input and tries to perform the optimal action considering the future steps with RL. The proposed MFMS-SFN system differs with the SFMS-SFN by introduce the LSTM network to encode the multi-frame historical heatmaps as the state for RL.
\end{itemize}

The spiral search and E2E Force-RL are force-based methods which can be applied to the real world directly without sim-to-real transfer. For the other vision-based methods, we conduct the same sim-to-real technique with our proposed system, which learns a seam segmentation module on the scene. We conduct efficient seam segmentation training with automatic data collection and annotation in the real world platform, where we collect 200 images with labelled masks as annotations by randomly change the robot-peg pose within 10 minutes. We use U-Net as backbone for segmentation and the training converges with 10 epochs in around 5 minutes. We achieve nearly perfection precision for segmentation with $MeanIoU$ \cite{c23} exceeding 0.97. The segmented image is took as input by the policy trained in simulation for position and orientation alignment  to achieve sim-to-real.

\textbf{Performance evaluation:} As shown in Tab. \ref{table_2}, the force-based methods like spiral search and E2E Force-RL can achieve high success rate and are general to the unseen geometries when the error is within 5mm. However, the methods are not efficient, and the success rate drops rapidly with the position error increasing to 10mm. Moreover,
in experiments we found it hard for the force-based methods to learn the orientation alignment especially with the yaw error in $z$-axis\cite{c6,c1}. Due to this reason, we take one-step back by performing 3-DoF peg-in-hole which only outputs translate displacements $dx, dy, dz$ in the action space as starred in Tab. \ref{table_2}, which is unfair to the vision-based 4-DoF peg-in-hole that considers the extra $z$-axis rotation error. The E2E Vision-RL can achieve both high success rate and efficiency by overfitting to the seen geometries during training, but pays for the price for not general to the unseen geometries. Our proposed system can realize both success rate and efficiency while being general to unseen geometries. In addition, the SFSS-SFN-re struggles to be general to the shape 7 and shape 10, but the SFMS-SFN and the MFMS-SFN have addressed it by introducing RL for decision making. The MFMS-SFN achieves better success rate and efficiency than the SFMS-SFN when the position error is 10mm, we infer that encoding the historical frames has helped to reduce the uncertainty for decision making.

\begin{figure}[t]
\includegraphics[scale=0.35]{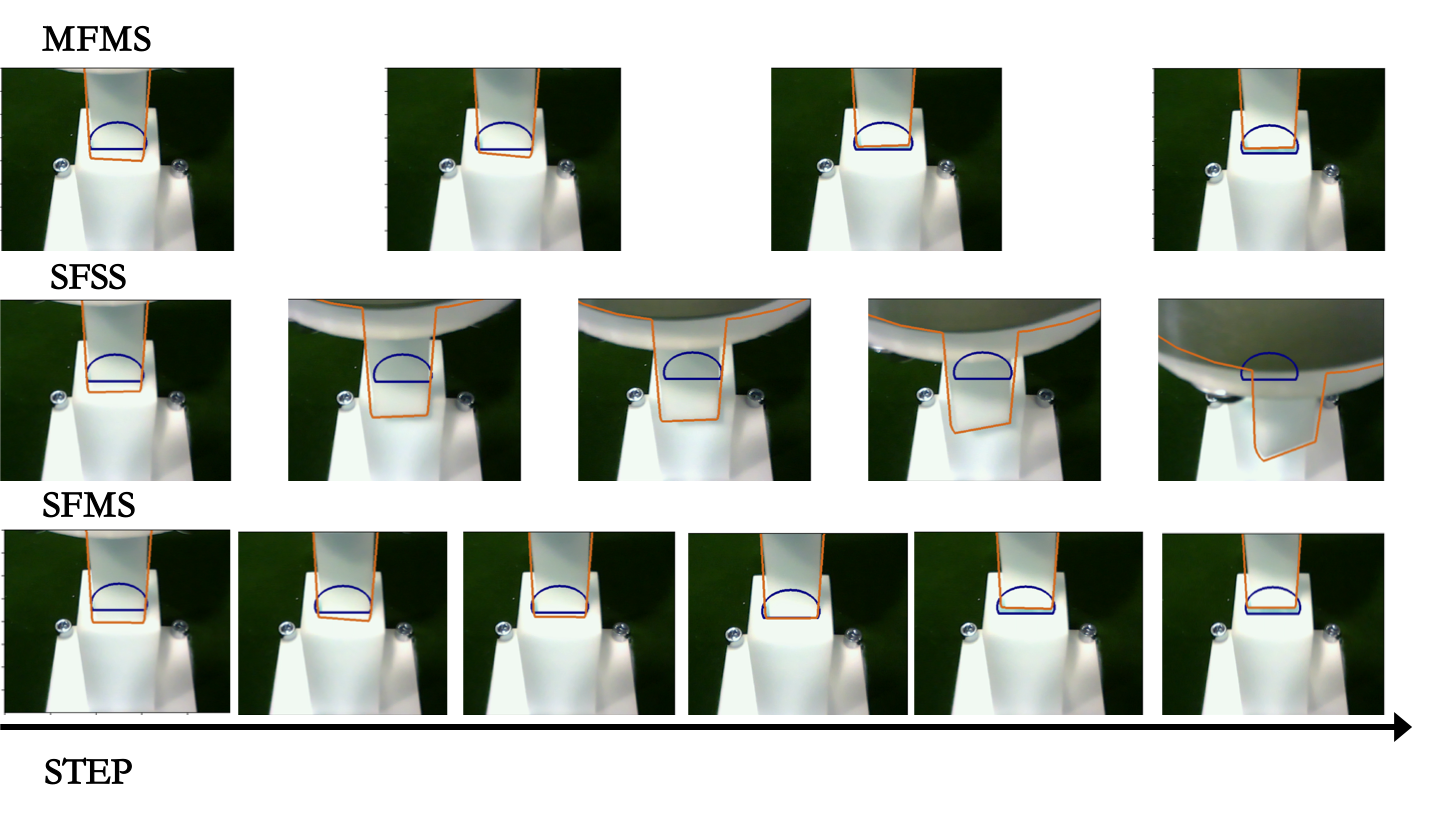}
\caption{MFMS performs fast and accurate alignment; SFSS drifts away the safe zone; SFMS achieves alignment with low efficiency.}
\label{Vision Occlusion}
\vspace{-0.5cm}
\end{figure}

\textbf{Vision occlusion case study:} We evaluate the robustness of our proposed system when the vision occlusion occurs at the initial states as shown in Fig. \ref{Vision Occlusion}. The SFSS fails to perform correct estimation and drifts off the safe zone where no historical or future information can help the policy to pull back. However, both the SFMS and the MFMS can recover from the visual blind but the MFMS is more efficient.

\section{Conclusions}
The paper presented a vision guided framework that achieves the peg-in-hole insertion with unseen shape generalization. On top of that, we improve the performance of the task with the insertion efficiency, robustness, and tight tolerance over the unseen shapes. With the help of the proposed sim-to-real module, we only need to retrain the segmentation module on the scene with automatic data collection and annotation, which is efficient in real world without human intervention. In the future work, we plan to pursue the tight insertion task with higher precision by combining both the vision and  force data.

\addtolength{\textheight}{-11cm}   









\end{document}